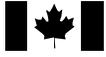

# *Unsupervised Learning of Semantic Orientation from a Hundred-Billion-Word Corpus*

P.D. Turney, National Research Council Canada

M.L. Littman, Stowe Research

May 16, 2002

# Unsupervised Learning of Semantic Orientation from a Hundred-Billion-Word Corpus

P.D. Turney, National Research Council Canada

M.L. Littman, Stowe Research

May 16, 2002

## Contents

## Abstract

The evaluative character of a word is called its *semantic orientation*. A positive semantic orientation implies desirability (e.g., "honest", "intrepid") and a negative semantic orientation implies undesirability (e.g., "disturbing", "superfluous"). This paper introduces a simple algorithm for unsupervised learning of semantic orientation from extremely large corpora. The method involves issuing queries to a Web search engine and using pointwise mutual information to analyse the results. The algorithm is empirically evaluated using a training corpus of approximately one hundred billion words — the subset of the Web that is indexed by the chosen search engine. Tested with 3,596 words (1,614 positive and 1,982 negative), the algorithm attains an accuracy of 80%. The 3,596 test words include adjectives, adverbs, nouns, and verbs. The accuracy is comparable with the results achieved by Hatzivassiloglou and McKeown (1997), using a complex four-stage supervised learning algorithm that is restricted to determining the semantic orientation of adjectives.

## Introduction

Many words communicate the speaker's evaluation of the item that is under discussion as desirable or undesirable. This evaluative character a word is called its *semantic orientation*. A word with a positive semantic orientation conveys the evaluation that the item is desirable (e.g., "beautiful") and a negative orientation conveys the evaluation that the item is undesirable (e.g., "absurd").

This paper presents a general strategy for inferring semantic orientation from semantic association. Section 1 gives two examples of this strategy, one based on mutual information (Church & Hanks, 1989) and the other based on Latent Semantic Analysis (Landauer and Dumais, 1997).

Related work is examined in Section 2. Hatzivassiloglou and McKeown (1997) have developed a supervised learning algorithm that learns semantic orientation from linguistic constraints on the use of adjectives in conjunctions.

The experimental results are presented in Section 3. The algorithms are evaluated using 3,596 words (1,614 positive and 1,982 negative) taken from the General Inquirer lexicon (Stone *et al.*, 1966). These words include adjectives, adverbs, nouns, and verbs. An accuracy of 80% is attained, using an unlabeled training corpus of approximately one hundred billion words. The interpretation of the experimental results is given in Section 4.

Section 5 lists some potential applications of the algorithms, such as filtering "flames" (abusive messages) for newsgroups (Spertus, 1997) and tracking opinions in on-line discussions (Tong, 2001). The paper concludes with some speculation about possible extensions of this approach to other tasks.

## 1. Semantic Orientation from Association

The general strategy in this paper is to infer semantic orientation from semantic association. Seven positive words (good, nice, excellent, positive, fortunate, correct, and superior) and seven negative words (bad, nasty, poor, negative, unfortunate, wrong, and inferior) are used as paradigms of positive and negative semantic orientation. The semantic orientation of a given word is calculated from the strength of its association with the seven positive words, minus the strength of its association with the seven negative words. These fourteen words were chosen using intuition. They are based on opposing pairs (good/bad, nice/nasty, excellent/poor, etc.).

It could be argued that this is a supervised learning algorithm with fourteen labeled training examples and millions or billions of unlabeled training examples. However, it seems more appropriate to say that the paradigm words are defining semantic orientation, rather than training the algorithm.

This general strategy is called SO-A (Semantic Orientation from Association). Selecting particular measures of word association results in particular instances of the strategy. This paper examines SO-PMI-IR (Semantic Orientation from Pointwise Mutual Information and Information Retrieval) and SO-LSA (Semantic Orientation from Latent Semantic Analysis).

### 1.1 Semantic Orientation from PMI-IR

PMI-IR (Turney, 2001) uses Pointwise Mutual Information (PMI) to calculate the strength of the semantic association between words (Church & Hanks, 1989). Word co-occurrence statistics are obtained using Information Retrieval (IR). PMI-IR has been empirically evaluated using 80 synonym test questions from the Test of English as a Foreign Language (TOEFL), obtaining a score of 74% (Turney, 2001). For comparison, Latent Semantic Analysis (LSA) attains a score of 64% on the same 80 TOEFL questions (Landauer & Dumais, 1997).

The Pointwise Mutual Information (PMI) between two words, $word_1$ and $word_2$, is defined as follows (Church & Hanks, 1989):

$$\text{PMI}(word_1, word_2) \;=\; \log_2\!\left(\frac{\text{p}(word_1 \; \& \; word_2)}{\text{p}(word_1) \; \text{p}(word_2)}\right) \tag{1}$$

Here, $\text{p}(word_1 \; \& \; word_2)$ is the probability that $word_1$ and $word_2$ co-occur. If the words are statistically independent, the probability that they co-occur is given by the product $\text{p}(word_1) \; \text{p}(word_2)$. The ratio between $\text{p}(word_1 \; \& \; word_2)$ and $\text{p}(word_1) \; \text{p}(word_2)$ is a measure of the degree of statistical dependence between the words. The log of the ratio is the amount of information that we acquire about the presence of one word when we observe the other.

The semantic orientation of a word, *word*, is calculated by SO-PMI-IR as follows:

$$\begin{aligned}\text{SO-PMI-IR}(word) = {} & \text{PMI}(word, \{positive\ paradigms\}) \\ & - \text{PMI}(word, \{negative\ paradigms\})\end{aligned} \tag{2}$$

In this equation, $\{positive\ paradigms\}$ represents the seven positive words (good, nice, excellent, positive, fortunate, correct, and superior) and $\{negative\ paradigms\}$ represents the seven negative words (bad, nasty, poor, negative, unfortunate, wrong, and inferior).

PMI-IR estimates PMI by issuing queries to a search engine (hence the IR in PMI-IR) and noting the number of hits (matching documents). The following experiments use the AltaVista Advanced Search engine[1], which indexes approximately 350 million Web pages (counting only those pages that are in English). Given a (conservative) estimate of 300 words per Web page, this represents a corpus of at least one hundred billion words.

AltaVista was chosen because it has a NEAR operator. The AltaVista NEAR operator constrains the search to documents that contain the words within ten words of one another, in either order. Previous work has shown that NEAR performs better than AND when measuring the strength of semantic association between words (Turney, 2001).

Let hits(*query*) be the number of hits returned, given the query *query*. The following equation can be derived from equations (1) and (2) with some minor algebraic manipulation, if co-occurrence is interpreted as NEAR:

$$\text{SO-PMI-IR}(word) \;=\; \log_2\!\left(\frac{\text{hits}(word \;\; \text{NEAR} \;\; p\_query) \; \text{hits}(n\_query)}{\text{hits}(word \;\; \text{NEAR} \;\; n\_query) \; \text{hits}(p\_query)}\right) \tag{3}$$

$$p\_query \;=\; (\text{good OR nice OR ... OR superior}) \tag{4}$$

$$n\_query \;=\; (\text{bad OR nasty OR ... OR inferior}) \tag{5}$$

Thus calculating the semantic orientation of a word requires four queries to AltaVista. Since hits($n\_query$) and hits($p\_query$) only need to be calculated once, the experiments required an average of only two queries per word. To avoid division by zero, 0.01 was added to the number of hits.[2]

A word, *word*, is classified as having a positive semantic orientation when SO-PMI-IR(*word*) is positive and a negative orientation when SO-PMI-IR(*word*) is negative. The magnitude of SO-PMI-IR(*word*) can be considered as the strength of the semantic orienta-tion.

---

1. See http://www.altavista.com/sites/search/adv.
2. The number 0.01 was arbitrarily chosen. This is a form of Laplace smoothing.

## 1.2   Semantic Orientation from LSA

SO-LSA applies Latent Semantic Analysis (LSA) to calculate the strength of the semantic association between words (Landauer and Dumais, 1997). LSA uses the Singular Value Decomposition (SVD) to analyze the statistical relationships among words in a corpus.

The first step is to use the text to construct a matrix $\mathbf{A}$, in which the row vectors represent words and the column vectors represent chunks of text (e.g., sentences, paragraphs, documents). Each cell represents the *weight* of the corresponding word in the corresponding chunk of text. The *weight* is typically the TF.IDF score (Term Frequency times Inverse Document Frequency) for the word in the chunk. (TF.IDF is a standard tool in Information Retrieval.)

The next step is to apply SVD to $\mathbf{A}$, to decompose $\mathbf{A}$ into a product of three matrices $\mathbf{U}\Sigma\mathbf{V}^T$, where $\mathbf{U}$ and $\mathbf{V}$ are in column orthonormal form (i.e., the columns are orthogonal and have unit length) and $\mathbf{S}$ is a diagonal matrix of *singular values* (hence SVD). If $\mathbf{A}$ is of rank $r$, then $\Sigma$ is also of rank $r$. Let $\Sigma_k$, where $k < r$, be the matrix produced by removing from $\Sigma$ the $r$ - $k$ columns and rows with the smallest singular values, and let $\mathbf{U}_k$ and $\mathbf{V}_k$ be the matrices produced by removing the corresponding columns from $\mathbf{U}$ and $\mathbf{V}$. The matrix $\mathbf{U}_k\Sigma_k\mathbf{V}_k^T$ is the matrix of rank $k$ that best approximates the original matrix $\mathbf{A}$, in the sense that it minimizes the sum of the squares of the approximation errors. We may think of this matrix $\mathbf{U}_k\Sigma_k\mathbf{V}_k^T$ as a "smoothed" or "compressed" version of the original matrix $\mathbf{A}$.

SVD may be viewed as a form of principal components analysis. LSA works by measuring the similarity of words using this compressed matrix, instead of the original matrix. The similarity of two words, $\text{LSA}(word_1, word_2)$, is measured by the cosine of the angle between their corresponding compressed row vectors.

The semantic orientation of a word, *word*, is calculated by SO-LSA as follows:

$$\text{SO-LSA-IR}(word) = \text{LSA}(word, \{positive\ paradigms\})$$
$$- \text{LSA}(word, \{negative\ paradigms\}) \tag{6}$$

$$= [\text{LSA}(word, \text{good}) + \ldots + \text{LSA}(word, \text{superior})]$$
$$- [\text{LSA}(word, \text{bad}) + \ldots + \text{LSA}(word, \text{inferior})] \tag{7}$$

As with SO-PMI-IR, a word, word, is classified as having a positive semantic orientation when SO-LSA(*word*) is positive and a negative orientation when SO-LSA(*word*) is negative. The magnitude of SO-LSA(*word*) represents the strength of the semantic orientation.

## 2.   *Related Work*

This work is most closely related to Hatzivassiloglou and McKeown's (1997) work on predicting the semantic orientation of adjectives. They note that there are linguistic constraints on the semantic orientations of adjectives in conjunctions. As an example, they present the following three sentences:

1. The tax proposal was simple and well-received by the public.
2. The tax proposal was simplistic but well-received by the public.
3. (*) The tax proposal  was simplistic and well-received by the public.

The third sentence is incorrect, because we use "and" with adjectives that have the same seman-tic orientation ("simple" and "well-received" are both positive), but we use "but" with adjectives that have different semantic orientations ("simplistic" is negative).

Hatzivassiloglou and McKeown (1997) use a four-step supervised learning algorithm to infer the semantic orientation of adjectives from constraints on conjunctions:

1. All conjunctions of adjectives are extracted from the given corpus.

2. A supervised learning algorithm combines multiple sources of evidence to label pairs of adjectives as having the same semantic orientation or different semantic orientations. The result is a graph where the nodes are adjectives and links indicate sameness or difference of semantic orientation.

3. A clustering algorithm processes the graph structure to produce two subsets of adjectives, such that links across the two subsets are mainly different-orientation links, and links inside a subset are mainly same-orientation links.

4. Since it is known that positive adjectives tend to be used more frequently than negative adjectives, the cluster with the higher average frequency is classified as having positive semantic orientation.

For brevity, we will call this the HM algorithm.

Like SO-PMI-IR and SO-LSA, HM can produce a real-valued number that indicates both the direction (positive or negative) and the strength of the semantic orientation. The clustering algorithm (step 3 above) can produce a "goodness-of-fit" measure that indicates how well an adjective fits in its assigned cluster.

Hatzivassiloglou and McKeown (1997) used a corpus of 21 million words and evaluated HM with 1,336 manually-labeled adjectives (657 positive and 679 negative). Their results are given in Table 1.[3] HM classifies adjectives with accuracies ranging from 78% to 92%, depending on the Alpha parameter, described next.

Table 1: The accuracy of HM with a 21 million-word corpus.

| Alpha threshold | Accuracy | Size of test set | Percent of "full" test set |
| --- | --- | --- | --- |
| 2 | 78.08% | 730 | 100.0% |
| 3 | 82.56% | 516 | 70.7% |
| 4 | 87.26% | 369 | 50.5% |
| 5 | 92.37% | 236 | 32.3% |

For a given adjective, Alpha is a measure of the confidence that the adjective will be correctly classified by HM, given the amount of data about the adjective that is available in the corpus. A threshold on Alpha is used to partition the 1,336 labeled adjectives into training and testing sets. For example, the first row in the table shows the accuracy when the training set is 606 adjectives with Alpha equal 1 and the testing set is 730 adjectives with Alpha greater than or equal to 2. As expected, the accuracy rises as the more difficult adjectives (adjectives with low Alpha) are moved from the testing set into the training set.

This algorithm is able to achieve good accuracy levels, but it has some limitations. In contrast with learning semantic orientation from semantic association, HM is restricted to adjectives, it requires labeled adjectives as training data, and the four-step process is difficult to implement and to analyze theoretically.

---

3. This table is derived from Table 3 in Hatzivassiloglou and McKeown (1997).

## 3. *Experiments*

The following experiments use the General Inquirer lexicon (Stone *et al.*, 1966) as a benchmark to evaluate the learning algorithms.[4] This lexicon has 182 categories of word tags and 11,788 words. The words tagged "Positiv" (1,915 words) and "Negativ" (2,291 words) have (respectively) positive and negative semantic orientations. Words with multiple senses may have multiple entries in the lexicon. The list of 3,596 words (1,614 positive and 1,982 negative) used in the subsequent experiments was generated by reducing multiple-entry words to single entries. Table 2 lists some examples.

Table 2: Examples of "Positiv" and "Negativ" words.

| Positiv | | Negativ | |
|---------|---------|-------------|----------|
| abide | absolve | abandon | abhor |
| ability | absorbent | abandonment | abject |
| able | absorption | abate | abnormal |
| abound | abundance | abdicate | abolish |

### 3.1 Experiments with SO-PMI-IR

Table 3 shows the accuracy of SO-PMI-IR with a training corpus of 350 million Web pages (at least one hundred billion words). These are the English Web pages that are indexed by AltaVista. In this table, strength of the semantic orientation was used as a measure of confidence that the word will be correctly classified. Test words were sorted in descending order of the absolute value of their semantic orientation and the top ranked words (the highest confidence words) were then classified. For example, the second row in Table 3 shows the accuracy when the top 75% were classified and the bottom 25% (with lowest confidence) were ignored. .

Table 3: The accuracy of SO-PMI-IR with a one-hundred-billion word corpus.

| Percent of full test set | Size of test set | Accuracy |
|--------------------------|------------------|----------|
| 100% | 3596 | 79.70% |
| 75% | 2697 | 86.43% |
| 50% | 1798 | 90.04% |
| 25% | 899 | 92.21% |

Although Hatzivassiloglou and McKeown's (1997) experiment is different in several ways from this experiment (different algorithms, different test words, different corpora), Tables 1 and 3 show very similar results. With 100% of the test set words, HM has an accuracy of 78% and SO-PMI-IR has an accuracy of 80%. With 50% of the (higher confidence) test set words, the corresponding accuracies are 87% and 90%.

Table 4 shows the accuracy with 7 million Web pages (at least two billion words). This reduced corpus was produced by adding "AND host:.ca" to every query, which restricts the search results to the Web pages in the Internet domain "ca" (Canada).

Although the corpus in Table 4 is only 2% of the size of the corpus in Table 3, the accuracy drops only slightly, less than 3%. Interestingly, the gap between Table 1 (HM) and Table 4 is very small.

---

4. The General Inquirer lexicon is available for researchers at http://www.wjh.harvard.edu/~inquirer/.

Table 4: The accuracy of SO-PMI-IR with a two-billion word corpus.

| Percent of full test set | Size of test set | Accuracy |
|:---:|:---:|:---:|
| 100% | 3596 | 77.47% |
| 75% | 2697 | 83.98% |
| 50% | 1798 | 87.76% |
| 25% | 899 | 89.43% |

Table 5 shows the accuracy of SO-PMI-IR with a ten-million word corpus, the TASA-ALL corpus that has been used to train LSA.[5] This corpus is 0.5% of the size of the corpus in Table 4, and there is now a large drop in accuracy. The accuracy also does not increase smoothly as we decrease the percentage of the test set that is classified, which shows that the magnitude of the semantic orientation is no longer a good indicator of the confidence in the classification of a word.[6]

Table 5: The accuracy of SO-PMI-IR with a ten-million word corpus.

| Percent of full test set | Size of test set | Accuracy |
|:---:|:---:|:---:|
| 100% | 3596 | 62.32% |
| 75% | 2697 | 63.40% |
| 50% | 1798 | 54.67% |
| 25% | 899 | 65.07% |

The TASA-ALL corpus is not indexed by AltaVista. The above results were generated by emulating AltaVista on a local copy of the TASA-ALL corpus.

## 3.2 Experiments with SO-LSA

Table 6 shows the accuracy of SO-LSA with a 37,651 document corpus (approximately ten million words). This experiment used the online demonstration of LSA[7] with the TASA-ALL corpus. The corpus was used to generate a matrix A with 92,409 rows (words) and 37,651 columns (chunks of text) and SVD reduced the matrix to 300 dimensions. In the online demonstration, this is called the "General Reading up to 1st year college (300 factors)" topic space.

The results in Table 6 can be directly compared with the results in Table 5, since they are based on the same corpus. SO-PMI-IR and SO-LSA have approximately the same accuracy when evaluated on the full test set, but SO-LSA rapidly pulls ahead as we decrease the percentage of the test set that is classified. It appears that the magnitude of SO is a better indicator of confidence for SO-LSA than for SO-PMI-IR, at least when the corpus is relatively small.

---

5. See http://lsa.colorado.edu/spaces.html.

6. It might be possible to improve these results by optimizing the Laplace smoothing factor (0.01), but we have not yet tried this.

7. See http://lsa.colorado.edu/.

Table 6: The accuracy of SO-LSA with a ten-million word corpus.

| Percent of full test set | Size of test set | Accuracy |
| --- | --- | --- |
| 100% | 3596 | 65.24% |
| 75% | 2697 | 71.04% |
| 50% | 1798 | 75.36% |
| 25% | 899 | 81.65% |

# 4. *Discussion of Results*

The interpretation of the results is complicated by the use of different test words (in Section 2 versus Section 3) and different corpora. Unfortunately, Hatzivassiloglou and McKeown's (1997) set of 1,336 manually-labeled adjectives was not available for testing SO-PMI-IR and SO-LSA. Also, LSA has not yet been scaled up to corpora of the sizes that are available for PMI-IR, so we were unable to evaluate SO-LSA on the larger corpora that were used to evaluate SO-PMI-IR (in Tables 3 and 4).

With these caveats in mind, the experiments suggest that the SO-PMI-IR can reach the same level of accuracy as HM, given a sufficiently large corpus. The results also hint that SO-LSA is able to use data more efficiently than SO-PMI-IR, and SO-LSA might surpass the 80% accuracy attained by SO-PMI-IR, given a corpus of comparable size.

HM is restricted to adjectives, it requires labeled training data, and it is complex. SO-PMI-IR and SO-LSA overcome these three limitations, but they appear to require larger corpora to achieve good accuracy levels.

# 5. *Applications*

The motivation of Hatzivassiloglou and McKeown (1997) was to use semantic orientation as a component in a larger system, to automatically identify antonyms and distinguish near synonyms. Both synonyms and antonyms typically have strong semantic associations, but synonyms generally have the same semantic orientation, whereas antonyms have opposite orientations.

Another potential application is filtering "flames" for newsgroups (Spertus, 1997). There could be a threshold, such that a newsgroup message is held for verification by the human moderator when the semantic orientation of a word drops below the threshold.

Tong (2001) presents a system for generating *sentiment timelines*. This system tracks online discussions about movies and displays a plot of the number of positive sentiment and negative sentiment messages over time. Messages are classified by looking for specific phrases that indicate the sentiment of the author towards the movie. Tong's (2001) system could benefit from the use of a learning algorithm, instead of (or in addition to) a hand-built lexicon. Advertisers could track advertising campaigns, politicians could track public opinion, reporters could track public response to current events, and stock traders could track financial opinions.

A related application is the analysis of survey responses to open ended questions. Commercial tools for this task include TextSmart[8] (by SPSS) and Verbatim Blaster[9] (by StatPac). These tools can be used to plot word frequencies or cluster responses into categories, but they do not currently analyze semantic orientation.

---

8. See http://www.spss.com/textsmart/.
9. See http://www.statpac.com/content-analysis.htm.

We are currently experimenting with the use of semantic orientation to classify reviews (e.g., movie reviews, automobile reviews, travel reviews). The average semantic orientation of the words in a review may be an indicator of whether the review is positive or negative. Table 7 shows the average semantic orientation of sentences selected from reviews of banks, from the Epinions site.[10] Five of these six randomly selected sentences are classified correctly.

Table 7: The average semantic orientation of some sample sentences from reviews of banks.

| Positive Reviews | Average SO |
|---|---|
| 1. I love the local branch, however **communication** may break down if they have to go through head office. | 0.1414 |
| 2. Bank of America gets my business because of its **extensive** branch and ATM network. | 0.1226 |
| 3. This bank has exceeded my **expectations** for the last ten years. | 0.1690 |
| Negative Reviews | Average SO |
| 1. Do not bank here, their website is even **worse** than their actual locations. | -0.0766 |
| 2. Use Bank of America only if you like the feeling of a stranger's **warm**, sweaty hands in your pockets. | 0.1535 |
| 3. If you want **poor** customer service and to lose money to ridiculous charges, Bank of America is for you. | -0.1314 |

In Table 7, for each sentence, the word with the strongest semantic orientation has been marked in bold. These bold words dominate the average and largely determine the orientation of the sentence as a whole. In the sentence that is misclassified as positive, the system is misled by the sarcastic tone. The negative orientations of "stranger's" and "sweaty" were not enough to counter the strong positive orientation of "warm".

Another application is in an automated chat system (a *chatbot*), to help decide whether a positive or negative response is most appropriate. Characters in software games would appear more realistic if they responded to semantic orientation.

## *Conclusion*

This paper has presented a general strategy for learning semantic orientation from semantic association, SO-A. Two instances of this strategy have been empirically evaluated, SO-PMI-IR and SO-LSA. The accuracy of SO-PMI-IR is comparable to the accuracy of HM, the algorithm of Hatzivassiloglou and McKeown (1997). SO-PMI-IR requires a large corpus, but it is simple, easy to implement, unsupervised, and it is not restricted to adjectives.

In the future, corpora of a hundred billion words will be common and the average desktop computer will be able to process them easily. Today, we can indirectly work with corpora of this size through Web search engines, as we have done in this paper. With a little bit of creativity, a Web search engine can tell us a lot about language use.

More specifically, the ideas in SO-A can likely be extended to many other semantic aspects of words. The General Inquirer lexicon has 182 categories of word tags and this paper has only used two of them (Stone *et al.*, 1966), so there is no shortage of future work.

---

10. See http://www.epinions.com/.